\documentclass[pdflatex,sn-mathphys-num]{sn-jnl}
\usepackage{graphicx}%
\usepackage{multirow}%
\usepackage{amsmath,amssymb,amsfonts}%
\usepackage{amsthm}%
\usepackage{mathrsfs}%
\usepackage[title]{appendix}%
\usepackage{xcolor}%
\usepackage{textcomp}%
\usepackage{manyfoot}%
\usepackage{booktabs}%
\usepackage{algorithm}%
\usepackage{algorithmicx}%
\usepackage{algpseudocode}%
\usepackage{listings}%
\theoremstyle{thmstyleone}%
\theoremstyle{thmstyletwo}%

\theoremstyle{thmstylethree}%

\begin{document}

\title[Article Title]{ARIADNE: A Perception-Reasoning Synergy Framework for Trustworthy Coronary Angiography Analysis}

\author[1]{\fnm{Zhan} \sur{Jin}}\email{zjin@stu.ouc.edu.cn}
\equalcont{These authors contributed equally to this work.}

\author[1]{\fnm{Yu} \sur{Luo}}\email{luoyu@stu.ouc.edu.cn}
\equalcont{These authors contributed equally to this work.}

\author[1]{\fnm{Yizhou} \sur{Zhang}}\email{zyz6596@stu.ouc.edu.cn}
\equalcont{These authors contributed equally to this work.}

\author[1]{\fnm{Ziyang} \sur{Cui}}\email{3196148390@qq.com}
\equalcont{These authors contributed equally to this work.}

\author[1]{\fnm{Yuqing} \sur{Wei}}\email{2659799366@qq.com}

\author[1]{\fnm{Xianchao} \sur{Liu}}\email{2486063350@qq.com}

\author*[1]{\fnm{Xueying} \sur{Zeng}}\email{zxying@ouc.edu.cn}

\author*[2]{\fnm{Qing} \sur{Zhang}}\email{qingzhang2019@foxmail.com}

\affil[1]{\orgdiv{School of Mathematical Sciences}, \orgname{Ocean University of China}, \orgaddress{\city{Qingdao}, \postcode{266100}, \state{Shandong}, \country{China}}}

\affil[2]{\orgdiv{Department of Cardiology, Qilu Hospital (Qingdao), Cheeloo College of Medicine}, \orgname{Shandong University}, \orgaddress{\street{No. 758 Hefei Road}, \city{Qingdao}, \postcode{266000}, \state{Shandong}, \country{China}}}

\abstract{Conventional pixel-wise loss functions fail to enforce topological constraints in coronary vessel segmentation, producing fragmented vascular trees despite high pixel-level accuracy. We present ARIADNE, a two-stage framework coupling preference-aligned perception with RL-based diagnostic reasoning for topologically coherent stenosis detection. The perception module employs DPO to fine-tune the Sa2VA vision-language foundation model using Betti number constraints as preference signals, aligning the policy toward geometrically complete vessel structures rather than pixel-wise overlap metrics. The reasoning module formulates stenosis localization as a Markov Decision Process with an explicit rejection mechanism that autonomously defers ambiguous anatomical candidates such as bifurcations and vessel crossings, shifting from coverage maximization to reliability optimization. On 1,400 clinical angiograms, ARIADNE achieves state-of-the-art centerline Dice of 0.838, reduces false positives by 41\% compared to geometric baselines. External validation on multi-center benchmarks ARCADE and XCAD confirms generalization across acquisition protocols. This represents the first application of DPO for topological alignment in medical imaging, demonstrating that preference-based learning over structural constraints mitigates topological violations while maintaining diagnostic sensitivity in interventional cardiology workflows. The code is available at https://github.com/qimingfan10/ARIADNE.}

\keywords{Coronary Angiography, Foundation Models, Direct Preference Optimization, Reinforcement Learning, Topological Consistency, Stenosis Detection}



\maketitle

\section{Introduction}\label{sec1}
Coronary Artery Disease (CAD) remains a leading cause of morbidity and mortality worldwide\cite{GBD2023}, requiring diagnostic modalities that provide accurate, reproducible, and efficient assessment. Invasive X-ray Coronary Angiography (XCA) serves as the primary tool for CAD diagnosis and guidance of Percutaneous Coronary Interventions (PCI)\cite{Jennifer2022}, offering high temporal resolution necessary for visualizing hemodynamic flow\cite{Ramos2025lightweight}. However, current clinical workflows rely heavily on manual interpretation, a process characterized by significant inter-observer variability and susceptibility to clinician fatigue\cite{Miguel2022}. As healthcare institutions universally adopt Picture Archiving and Communication Systems (PACS), a critical gap persists between passive image storage and active, automated clinical interpretation. While hospitals have implemented digital image storage, they lack automated systems capable of transforming raw imaging data into actionable clinical insights. The growing volume of interventional procedures makes purely manual interpretation increasingly unsustainable, creating demand for Computer-Aided Diagnosis systems that can bridge the gap between data acquisition and clinical decision-making.

Accurate segmentation of the coronary vascular tree represents a fundamental prerequisite for automated coronary analysis. Over the past decade, Convolutional Neural Networks (CNNs), particularly U-Net\cite{Ronneberger2015unet} and its attention-enhanced variants such as CS-Net and SA-UNet\cite{Li2024csnet,Wang2022saunet}, have dominated the field. More recently, Vision Transformers (ViTs) have been introduced to capture global spatial relationships\cite{Chen2021TransUNet}. Despite achieving high pixel-level performance metrics, these models face a critical limitation in preserving vascular topology. Traditional loss functions, including Cross-Entropy and Dice Loss\cite{Milletari2016VNet}, optimize pixel-level accuracy independently without explicitly penalizing topological errors\cite{Shit2021CVPR}. Consequently, these models frequently produce fragmented vessel trees where distal branches appear disconnected, particularly due to signal loss in thin vessels during downsampling operations\cite{Chang2024optimizing}. In coronary hemodynamics analysis, topological connectivity is essential; a segmentation with high Dice score remains insufficient for clinical use if discontinuities prevent accurate centerline extraction and subsequent geometric analysis.

The recent emergence of foundation-scale Vision-Language Models (VLMs) has introduced a complementary approach to medical image segmentation. Models such as SAM3\cite{Carion2025SAM3} and MedSAM3\cite{Liu2025MedSAM3} leverage large language models to enable prompt-based segmentation, where textual descriptions guide mask generation. These architectures demonstrate impressive semantic understanding, correctly identifying what constitutes a vessel based on learned visual-linguistic correspondences. However, their training on generic natural image datasets creates a fundamental semantic-topological gap: while VLMs comprehend the conceptual category of a vascular structure, they lack the domain-specific anatomical priors necessary to enforce structural continuity in low-contrast, projection-based X-ray angiography. Empirical evaluation reveals that general-purpose VLMs consistently produce semantically correct but topologically fragmented segmentations—correctly classifying pixels as vessel while failing to maintain the connected tree structure essential for hemodynamic modeling. This failure stems from their optimization objective: VLMs maximize pixel-level overlap (Dice, IoU\cite{Rezatofighi2019GIoU}) between predicted and ground-truth masks, a criterion that remains agnostic to whether the resulting mask forms a continuous vascular network or a collection of disconnected segments. In coronary angiography, where vessel diameters approach image resolution limits and contrast variability is substantial, the absence of explicit topological constraints results in high-confidence predictions of isolated vessel fragments that are clinically unusable for stenosis quantification or flow analysis.

This limitation in segmentation directly impacts the accuracy of stenosis detection systems. Current automated frameworks predominantly follow a sequential approach where segmentation and stenosis detection are performed as independent tasks\cite{Liu2023AQC,Huang2025}. In these systems, geometric algorithms traverse the segmented centerline to identify regions of narrowing. However, these deterministic algorithms lack the ability to distinguish pathological stenosis from common anatomical artifacts, including vessel crossings\cite{Hannink2014vesselness}, bifurcations, and foreshortening\cite{Yang2020CVPR}, resulting in elevated false positive rates. Conversely, while deep object detectors such as YOLO have been applied to direct lesion identification\cite{Diaz2025yolo}, they inherently lack the capacity to verify anatomical plausibility. Specifically, generic object detectors treat lesions as isolated bounding boxes, failing to validate whether a detected stenosis actually resides within a continuous, hemodynamically relevant vascular segment. These limitations have hindered clinical adoption due to the high rate of false alarms that reduce system reliability.

To address these fundamental challenges in coronary angiography automation, we propose ARIADNE (Anatomy-aware Reasoning for Integrated Angiography Diagnosis and Navigation Expert), a framework that bridges the gap between visual perception and clinical reasoning. Our central hypothesis is that robust diagnostic automation requires not only accurate visual recognition but also explicit alignment with the hierarchical reasoning patterns employed by expert clinicians. Building on recent advances in preference-based learning from the artificial intelligence community, we integrate Direct Preference Optimization (DPO)\cite{Rafailov2023dpo} with Reinforcement Learning (RL)\cite{Schulz2021RL} to create a two-stage diagnostic pipeline. In the perception stage, we apply DPO to fine-tune a vision-language foundation model (Sa2VA)\cite{Wallace2024CVPR,Konwer2025enhancing}, using comparative preferences derived from centerline continuity profiles, quantified via clDice\cite{Shit2021CVPR}, to guide the model toward structurally coherent vessel segmentations. Unlike general VLMs that optimize for semantic correctness through pixel overlap, our preference-based approach explicitly rewards topological continuity—teaching the model that a mask with 92\% Dice score but preserved connectivity is preferable to a 95\% Dice mask with fragmented branches. This enables the model to harness the semantic power of vision-language architectures while enforcing the geometric rigor required for hemodynamic analysis. To consolidate this topological reasoning against weak visual signals, we further incorporate a Hard Sample Focused Training (HSFT) strategy. By concentrating optimization resources on the most diagnostically uncertain subsets—such as complex bifurcations and distal vessels—this mechanism achieves significant computational efficiency while ensuring robust performance in anatomically challenging regions where global statistics often mask local failures. The resulting topologically coherent vessel trees provide a clinically valid foundation for the reasoning stage: a RL-based navigation agent that performs sequential decision-making for stenosis detection. Critically, this agent incorporates an explicit rejection mechanism\cite{Chow2003ro} that mirrors the clinical workflow where radiologists flag ambiguous cases for secondary review. By allowing the system to abstain from uncertain predictions, we shift the operational paradigm from maximizing coverage to maximizing reliability, thereby reducing false positive rates while maintaining high sensitivity for clear-cut lesions. This perception-to-reasoning architecture reflects the natural diagnostic workflow, where accurate anatomical reconstruction serves as the perceptual foundation for subsequent lesion localization and characterization.

This work makes three primary contributions to the automation of coronary angiography interpretation:
\begin{enumerate}
    \item \textbf{Perception Framework:} We introduce a preference-based optimization approach that aligns VLMs with topological constraints in vessel segmentation. By applying DPO to comparative vessel tree examples, our method achieves topologically consistent segmentations without requiring pixel-level annotation of connectivity features, augmented by a hard-sample mining strategy that enhances computational efficiency in complex anatomical scenarios.
    \item \textbf{Reasoning Algorithm:} We formulate stenosis detection as a sequential navigation task guided by RL, incorporating an explicit rejection mechanism that allows the system to defer ambiguous cases. This clinical workflow-aligned approach substantially reduces false positive rates in anatomically complex regions while maintaining high sensitivity for definitive lesions.
    \item \textbf{Clinical Validation:} We demonstrate that integrating topologically-aware perception with rejection-enabled reasoning achieves state-of-the-art diagnostic performance on standard coronary angiography benchmarks with a TPR of 0.867, supporting the hypothesis that anatomical validity is prerequisite to reliable automated diagnosis.
\end{enumerate}

\section{Methods}\label{sec2}

\subsection{Framework Overview}

\begin{figure}[htbp]
\centering
  \includegraphics[width=0.8\textwidth]{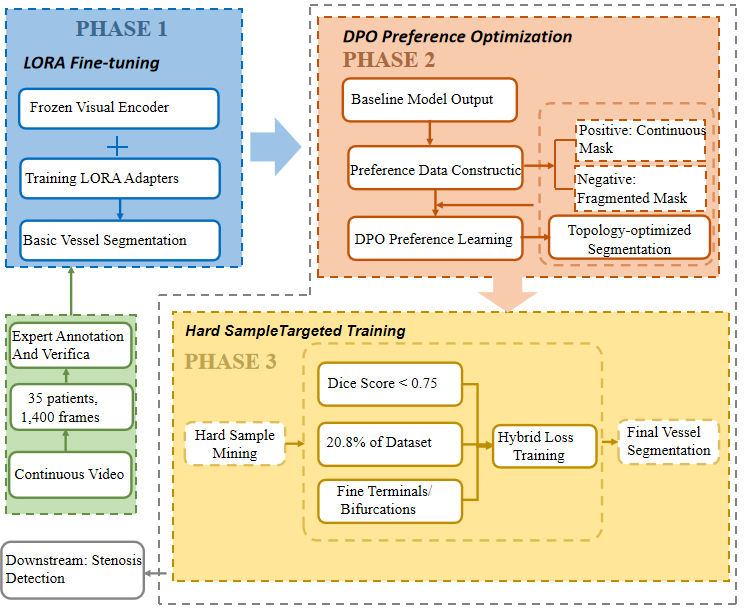}
  \caption{Training framework of Anatomy-Aware Segmentation}
  \label{fig:1}
\end{figure}

\begin{figure}[htbp]
\centering
  \includegraphics[width=\textwidth]{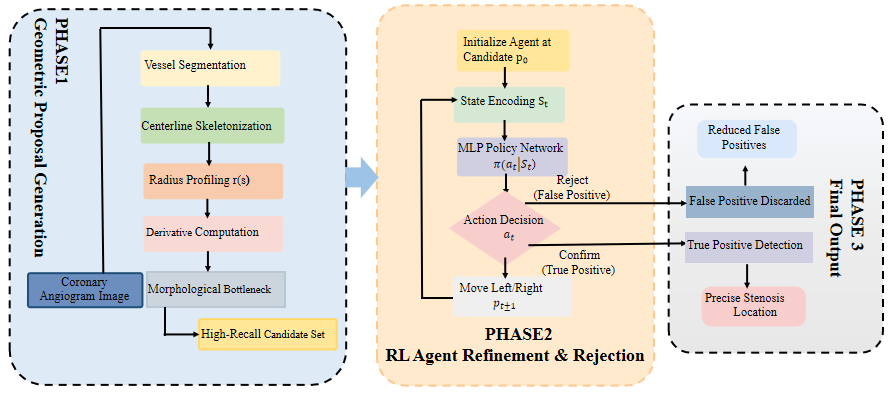}
  \caption{Training framework of Structure-Guided Reasoning}
  \label{fig:2}
\end{figure}

To operationalize the clinical requirement for topological continuity in angiographic analysis, the proposed ARIADNE framework is designed to emulate the hierarchical decision-making process of human experts. As illustrated in Fig.~\ref{fig:1} and Fig.~\ref{fig:2}, the system consists of two biomimetic stages that mirror the visual-cognitive workflow of expert interventional cardiologists: a perception module for anatomically consistent vascular reconstruction and a reasoning module for context-aware lesion localization.

The perception module employs the Sa2VA foundation model\cite{Yuan2025sa2va} with a progressive training strategy designed to enforce topological continuity throughout the segmentation process. We integrate DPO\cite{Rafailov2023dpo} into the training pipeline to align model outputs toward geometrically complete vessel structures rather than fragmented pixel-level predictions. This preference-based learning approach guides the model to preserve vascular connectivity without requiring exhaustive manual annotation of topological features, generating vessel masks that maintain the vascular continuity essential for downstream hemodynamic analysis. The resulting segmentation masks maintain structural integrity across vessel hierarchies, providing a clinically reliable anatomical scaffold for downstream diagnostic reasoning.

Building upon this topologically consistent representation, the reasoning module operates as a structure-guided diagnostic agent that navigates the extracted vessel skeleton to identify stenotic lesions. Rather than applying fixed statistical thresholds, we develop a RL agent that analyzes local geometric features—including radius gradients and curvature patterns—to perform context-aware lesion localization. Critically, the agent incorporates an explicit rejection mechanism to filter false positive detections arising from complex anatomical structures such as vessel crossings, bifurcations, and foreshortening artifacts. The effectiveness of this rejection mechanism is fundamentally dependent on the structural consistency provided by the perception module, demonstrating the essential interdependence between anatomical reconstruction and diagnostic decision-making within the ARIADNE framework.

\subsection{Anatomy-Aware Perception Module via Preference Alignment}

Coronary vessel segmentation requires bridging the semantic gap between low-level pixel intensities in fluoroscopic images and high-level anatomical knowledge of vascular topology. To achieve this integration, we employ the Sa2VA architecture\cite{Yuan2025sa2va}, a visual-language foundation model designed to align angiographic representations with structured clinical priors. Formally, given an input angiographic image $\mathbf{x} \in \mathbb{R}^{H \times W \times C}$, the architecture consists of three interdependent components that collectively enable topological alignment. The vision encoder $\mathcal{E}_v$, instantiated as InternViT-6B-448px\cite{Chen2024InternVL}, operates in a frozen state to extract robust high-semantic embeddings $\mathbf{z}_v = \mathcal{E}_v(\mathbf{x}) \in \mathbb{R}^{N \times d_v}$, where $N$ represents the number of patch tokens and $d_v$ denotes the embedding dimension. Rather than training exclusively on limited medical datasets, this generalized visual embedding space—pre-trained on large-scale natural images—provides stable feature representations across the varying contrast conditions and fluoroscopic noise characteristic of interventional cardiology. The language modeling component $\mathcal{L}$, based on InternLM2\cite{Cai2024internlm2}, maps anatomical directives (e.g., Segment the coronary artery) into a high-dimensional semantic space $\mathbf{z}_l = \mathcal{L}(\text{prompt}) \in \mathbb{R}^{d_l}$. This foundation model approach enables semantic representation of vascular terminology and supports potential generalization to complex clinical queries without architectural modification. Low-Rank Adaptation (LoRA)\cite{Hu2021lora} with rank $r = 16$ is applied to adapt pre-trained linguistic knowledge to coronary anatomy while avoiding full-parameter retraining costs. The integration between visual and linguistic streams forms the anatomical scaffold of the framework, where trainable projection layers align semantic embeddings with visual features to condition the SAM-2 mask decoder $\mathcal{D}$\cite{Ravi2024sam2}, yielding the segmentation mask $\mathbf{y} = \mathcal{D}(\mathbf{z}_v, \mathbf{z}_l) \in \{0, 1\}^{H \times W}$. By embedding clinical priors into the decoding process, this architecture suppresses background artifacts that possess similar pixel intensities to vessel structures but lack anatomical relevance, thereby focusing computational resources on topologically valid vascular components.

Traditional segmentation datasets consist of isolated static frames, creating ambiguity when anatomically complex structures or motion artifacts cannot be resolved without temporal context. To address this limitation, we implement a physiologically adaptive sampling strategy that leverages the continuous nature of angiographic video sequences. Rather than uniform temporal sampling—which introduces redundancy through near-duplicate frames while missing critical physiological events—we extract key frames that capture distinct hemodynamic states across the cardiac cycle. Frame selection explicitly targets systolic and diastolic phases to ensure exposure to the full range of vessel deformation, including lumen diameter variations and wall motion. Sampling additionally encompasses environmental variations in fluoroscopic imaging angles, X-ray intensity, and contrast bolus propagation phases, specifically arterial inflow, peak opacification, and venous washout. To maximize diagnostic relevance, we identify temporal hard clusters—consecutive sequences exhibiting low-confidence predictions that correspond to anatomically challenging regions such as distal vessel terminals, bifurcation zones, and overlapping vessel segments. This temporal mining strategy concentrates training samples on scenarios where visual ambiguity is maximal, mirroring the clinical workflow where radiologists dedicate additional scrutiny to diagnostically uncertain regions.

To evolve the model from a general-purpose segmentation system into a domain-specialized tool capable of preserving vascular topology, we implement a three-stage progressive training strategy that advances from basic visual pattern recognition to structured anatomical reasoning. In Stage 1, we establish visual pattern alignment through parameter-efficient transfer learning by freezing the vision encoder $\mathcal{E}_v$ to preserve generalized feature extraction capabilities while applying LoRA adapters to the InternLM2 language model and SAM-2 decoder. The model is trained on $N_1 = 1,220$ annotated angiogram samples by minimizing the Dice loss
\begin{equation}
    \mathcal{L}_{\text{Dice}} = 1 - \frac{2|\mathbf{y} \cap \mathbf{y}^*|}{|\mathbf{y}| + |\mathbf{y}^*|},
\end{equation}
where $\mathbf{y}^*$ denotes the ground truth mask. This initial alignment ensures the system can recognize vessel boundaries, tubular structures, and contrast-enhanced regions before addressing connectivity constraints. However, standard supervised objectives operate under pixel-wise independence assumptions—minimizing local discrepancies without explicitly penalizing topological violations such as vessel fragmentation.

Subsequently, to align the model with clinical reasoning principles that prioritize structural continuity, we incorporate topological constraints through DPO in Stage 2. Clinical validity is formally defined as topological connectivity: coronary arteries must form continuous tubular structures exhibiting $C^1$ continuity without artificial fragmentation, characterized by a single connected component—denoted by a Betti number of $\beta_0 = 1$—that preserves hemodynamic flow continuity. DPO enforces this constraint by formulating vascular connectivity as a preference learning problem where the objective is to maximize the likelihood margin between topologically valid and invalid segmentation states. Specifically, we construct a preference dataset $\mathcal{D}_{\text{pref}} = \{(\mathbf{x}^{(i)}, \mathbf{y}_w^{(i)}, \mathbf{y}_l^{(i)})\}_{i=1}^{N_2}$, where preference pairs are defined by adherence to topological constraints rather than pixel-wise overlap metrics. The preferred (winning) sample $\mathbf{y}_w$ is the ground truth segmentation, which satisfies global geometric constraints by exhibiting $\beta_0(\mathbf{y}_w) = 1$ and preserving flow continuity. The non-preferred (losing) sample $\mathbf{y}_l$ consists of hard negative examples mined from the Stage 1 policy $\pi_{\text{S1}}$—specifically, predictions with high pixel-level overlap $\text{Dice}(\mathbf{y}_l, \mathbf{y}^*) > 0.8$ but topological violations $\beta_0(\mathbf{y}_l) > \beta_0(\mathbf{y}_w)$, indicating vessel fragmentation or disjoint artifacts. The policy $\pi_\theta(\mathbf{y}|\mathbf{x})$, representing the probability distribution over segmentation masks, is optimized to assign higher probability to topologically connected samples over fragmented predictions through the DPO objective:
\begin{equation}
    \mathcal{L}_{\text{DPO}}(\pi_\theta; \pi_{\text{ref}}) = -\mathbb{E}_{(\mathbf{x}, \mathbf{y}_w, \mathbf{y}_l) \sim \mathcal{D}_{\text{pref}}} \left[ \log \sigma \left( \beta \log \frac{\pi_\theta(\mathbf{y}_w|\mathbf{x})}{\pi_{\text{ref}}(\mathbf{y}_w|\mathbf{x})} - \beta \log \frac{\pi_\theta(\mathbf{y}_l|\mathbf{x})}{\pi_{\text{ref}}(\mathbf{y}_l|\mathbf{x})} \right) \right],
\end{equation}
where $\pi_{\text{ref}}$ denotes the frozen Stage 1 policy serving as the reference model to prevent excessive deviation from learned visual features, $\beta = 0.1$ controls the KL-divergence penalty strength, and $\sigma$ represents the logistic sigmoid function. DPO optimizes the policy directly without training an explicit reward model, enabling efficient topological alignment through preference-based learning that guides the model toward geometrically complete vessel structures.

Finally, while DPO aligns the model with topological connectivity principles, performance remains inconsistent in anatomically complex scenarios where weak visual signals (low contrast, vessel overlap) destabilize the learned connectivity preference. To consolidate topological reasoning under diagnostic ambiguity, we implement HSFT in Stage 3 that concentrates computational resources on scenarios where clinical interpretation is most challenging. Rather than treating hard samples as statistical outliers, we identify temporal hard clusters—consecutive frames with Dice scores below threshold $\tau = 0.75$—which map to specific anatomical challenges: fine distal vessel terminals (diameter $< 1$mm), bifurcation zones where multiple branches diverge, and overlapping vessel segments in oblique projections. We define the hard sample subset $\mathcal{D}_{\text{hard}} = \{(\mathbf{x}, \mathbf{y}^*) \in \mathcal{D} \mid \text{Dice}(\pi_\theta(\mathbf{x}), \mathbf{y}^*) < \tau\}$, which constitutes 20.8\% of the dataset but accounts for the majority of topological errors. To enforce pixel-level accuracy in these regions while maintaining structural integrity, we apply the hybrid loss function
\begin{equation}
    \mathcal{L}_{\text{HSFT}} = \mathcal{L}_{\text{Dice}} + \lambda \mathcal{L}_{\text{BCE}},
\end{equation}
with $\lambda = 0.5$, where the Binary Cross-Entropy component
\begin{equation}
\mathcal{L}_{\text{BCE}} = -\sum_{i,j} \left[ \mathbf{y}^*_{ij} \log \mathbf{y}_{ij} + (1 - \mathbf{y}^*_{ij}) \log(1 - \mathbf{y}_{ij}) \right]
\end{equation}
provides pixel-wise gradients needed to refine vessel boundaries at bifurcations and terminals, while $\mathcal{L}_{\text{Dice}}$ maintains global structural consistency. This progressive strategy achieves 5$\times$ resource efficiency by focusing training on the most diagnostically relevant subset of samples, ensuring robust topological preservation across the full spectrum of anatomical complexity encountered in clinical practice.

\subsection{Structure-Guided Reasoning via Reinforcement Learning}

Building upon the topologically consistent vessel segmentations established by the perception module (Fig.~\ref{fig:1}), the reasoning stage (Fig.~\ref{fig:2}) translates structural information into diagnostic outputs by performing context-aware stenosis localization. This module leverages the topological integrity of DPO-enhanced segmentations to construct a navigable anatomical scaffold from which diagnostic candidates are systematically generated. Morphological thinning is applied to the refined binary segmentation mask $\mathbf{y} \in \{0,1\}^{H \times W}$ to extract the discrete vessel centerline $\mathcal{C} = \{p_1, p_2, \dots, p_N\}$ where $p_i \in \mathbb{R}^2$ denotes spatial coordinates, which serves as the navigation trajectory for subsequent geometric analysis. For each point $p_i$ on the skeleton, the local vessel radius $r(p_i)$ is computed using a Euclidean distance transform $\mathcal{D}(\mathbf{y})$, yielding $r(p_i) = \max\{d \mid B(p_i, d) \subset \mathbf{y}\}$ where $B(p_i, d)$ denotes a ball of radius $d$ centered at $p_i$. This generates a one-dimensional radius profile $r: [0, L] \to \mathbb{R}^+$ parameterized by arc length $s$ along the vessel's longitudinal axis. By analyzing this profile alongside its first and second derivatives $\nabla r(s) = \frac{dr}{ds}$ and $\nabla^2 r(s) = \frac{d^2r}{ds^2}$, we identify morphological bottlenecks as candidate locations $\mathcal{P}_{\text{cand}} = \{p \in \mathcal{C} \mid r(p) < \mu_r - k\sigma_r \land \nabla^2 r(p) > \theta_{\text{curv}}\}$, where $\mu_r$ and $\sigma_r$ denote the mean and standard deviation of the radius profile, $k = 1.5$ controls sensitivity, and $\theta_{\text{curv}}$ is a curvature threshold. This deterministic geometric process is deliberately configured for maximum sensitivity to generate a high-recall candidate set. However, this approach inherently produces false positives in anatomically complex regions such as bifurcations, vessel crossings, and natural tapering zones where geometric narrowing mimics pathological stenosis. These candidates therefore serve as initial proposals requiring subsequent verification, providing a high-recall, low-precision coordinate set that necessitates intelligent filtering through clinical reasoning.

To address the limitation of purely geometric detection, we formulate stenosis localization as a sequential decision-making process modeled as a Markov Decision Process (MDP). This formulation enables the system to perform context-aware diagnostic reasoning that distinguishes pathological stenoses from anatomical artifacts through analysis of local morphological patterns. Unlike static thresholding methods that apply fixed criteria uniformly across all vessel segments, RL allows the agent to adaptively evaluate each candidate based on its geometric neighborhood, mimicking the sequential visual inspection workflow employed by interventional cardiologists. The MDP is formally defined by the tuple $\mathcal{M} = (\mathcal{S}, \mathcal{A}, \mathcal{T}, \mathcal{R}, \gamma)$, where $\mathcal{S}$ denotes the state space encoding local vessel geometry, $\mathcal{A}$ represents the action space of navigational commands, $\mathcal{T}: \mathcal{S} \times \mathcal{A} \to \Delta(\mathcal{S})$ defines the state transition function, $\mathcal{R}: \mathcal{S} \times \mathcal{A} \to \mathbb{R}$ specifies the reward function encoding clinical priorities, and $\gamma \in [0,1)$ is the discount factor. The agent navigates the vascular skeleton to localize true stenoses while rejecting false alarms arising from benign anatomical variations.

Specifically, the state space $\mathcal{S} \subset \mathbb{R}^{16}$ encodes local morphological context at each candidate location. Each state vector $s_t = [r_{t-5:t+5}, \nabla r_{t-5:t+5}, Z_t, \kappa_t] \in \mathbb{R}^{16}$ comprises: (1) a normalized radius profile $r_{t-w:t+w}$ within a sliding window of half-width $w = 5$ centerline points, capturing the geometric progression of vessel lumen narrowing; (2) first-order derivatives $\nabla r_{t-w:t+w}$ quantifying morphological gradients to detect abrupt transitions characteristic of stenotic lesions; (3) local Z-score $Z_t = \frac{r(p_t) - \mu_r}{\sigma_r}$ benchmarking the degree of narrowing against the vessel's baseline caliber to distinguish significant stenoses from normal anatomical variation; and (4) local curvature $\kappa_t = \nabla^2 r(p_t)$ capturing geometric sharpness. The action space $\mathcal{A} = \{\text{Left}, \text{Right}, \text{Confirm}, \text{Reject}\}$ consists of discrete navigational commands, where lateral movements $\{\text{Left}, \text{Right}\}$ implement spatial translation $p_{t+1} = p_t \pm \Delta p$ with step size $\Delta p = 3$ pixels to enable fine positional adjustment toward the precise stenosis center. Critically, the Reject action implements an explicit abstention mechanism that transitions the agent to the next candidate in $\mathcal{P}_{\text{cand}}$ without issuing a diagnostic prediction, allowing autonomous dismissal of ambiguous candidates where local geometry superficially resembles pathology—such as bifurcation points where parent vessels exhibit natural narrowing as they split into smaller daughter branches. This rejection capability mirrors the clinical triage workflow where radiologists defer uncertain cases for secondary review rather than issuing potentially erroneous diagnoses, fundamentally shifting the operational paradigm from coverage maximization to reliability optimization and reducing false positive rates while maintaining high sensitivity for definitive lesions.

To align agent behavior with clinical diagnostic priorities, the reward function $\mathcal{R}: \mathcal{S} \times \mathcal{A} \to \mathbb{R}$ explicitly encodes the asymmetric costs of diagnostic errors, distinguishing between active detection failures, termed False Positives, and passive omission errors, or False Negatives. The reward function is formally structured to incentivize the correct rejection of anatomical artifacts while severely penalizing missed diagnoses:
\begin{equation}
R(s_t, a_t) =
\begin{cases}
r_{\text{TP}} & \text{if } a_t = \text{Confirm} \land \delta(p_t, G) \leq \tau \quad \text{(True Positive)} \\
r_{\text{FP}} & \text{if } a_t = \text{Confirm} \land \delta(p_t, G) > \tau \quad \text{(False Positive)} \\
r_{\text{TN}} & \text{if } a_t = \text{Reject} \land \delta(p_t, G) > \tau \quad \text{(Correct Rejection)} \\
r_{\text{FN}} & \text{if } a_t = \text{Reject} \land \delta(p_t, G) \leq \tau \quad \text{(False Negative)} \\
r_{\text{step}} & \text{otherwise}
\end{cases}
\end{equation}
where $\delta(p_t, G) = \|p_t - G\|_2$ represents the Euclidean distance to the nearest ground truth stenosis centroid $G$, and $\tau = 75$ pixels defines the localization tolerance. Hyperparameters are calibrated to reflect safety-critical clinical constraints: $r_{\text{TP}} = +50$ rewards accurate localization; $r_{\text{FP}} = -10$ penalizes false alarms to reduce clinician fatigue; $r_{\text{TN}} = +10$ explicitly rewards the agent for correctly identifying and rejecting ambiguous artifacts such as vessel crossings; and $r_{\text{FN}} = -50$ imposes a maximal penalty for rejecting a true stenosis, ensuring high sensitivity. A step cost $r_{\text{step}} = -1$ encourages efficient navigation. The optimal policy $\pi^*: \mathcal{S} \to \Delta(\mathcal{A})$ maximizes the expected cumulative discounted reward $\mathbb{E}_{\pi}\left[\sum_{t=0}^{\infty} \gamma^t R(s_t, a_t)\right]$ with discount factor $\gamma = 0.99$. Policy optimization is performed using Proximal Policy Optimization (PPO)\cite{Schulman2017PPO}, which ensures stable gradient updates by constraining the policy update through the clipped surrogate objective:
\begin{equation}
\mathcal{L}_{\text{PPO}}(\theta) = \mathbb{E}_t\left[\min\left(\rho_t(\theta)\hat{A}_t, \text{clip}(\rho_t(\theta), 1-\epsilon, 1+\epsilon)\hat{A}_t\right)\right]
\end{equation}
where $\rho_t(\theta) = \frac{\pi_\theta(a_t|s_t)}{\pi_{\theta_{\text{old}}}(a_t|s_t)}$ denotes the probability ratio between successive policy iterations, $\hat{A}_t$ is the generalized advantage estimate, and $\epsilon = 0.2$ defines the clipping range. This clipping mechanism prevents destructive updates that could destabilize the learned diagnostic strategy. The policy network $\pi_\theta$ employs a Multi-Layer Perceptron architecture with layer dimensions $[16 \to 256 \to 128 \to 64 \to |\mathcal{A}|]$ and ReLU activations, parameterized by weights $\theta \in \mathbb{R}^d$. Rather than employing recurrent architectures such as LSTMs or GRUs, this feedforward design enforces a strictly Markovian decision process where the policy $\pi_\theta(a_t|s_t)$ conditions exclusively on the current state $s_t$, ensuring diagnostic decisions remain invariant to the vessel's prior trajectory and maintaining computational efficiency with inference time $< 50$ms per candidate, suitable for real-time clinical deployment.

\section{Experiments}\label{sec3}
To validate the proposed perception-reasoning framework, the experimental design was structured to address two primary objectives: (1) evaluation of topological consistency in vascular segmentation across diverse angiographic conditions, and (2) assessment of stenosis detection accuracy and false positive management in anatomically complex scenarios.

\subsection{Datasets and Sampling Strategy}

The experimental foundation was established through a video-based acquisition strategy designed to capture morphological diversity in coronary angiography, supplemented by external validation on publicly available datasets to assess domain generalization.

A proprietary dataset was curated from coronary angiography video sequences acquired at Guizhou Aviation Industry Group 302 Hospital using a Siemens angiography system. The collection process utilized temporal information from video streams to ensure comprehensive representation of vessel morphology across 35 patients. The dataset comprises 1,400 high-resolution images at 512$\times$512 resolution from 35 patients, with an average of 40 frames extracted per patient to capture varying vessel angulations and contrast conditions. To prevent data leakage inherent in video-based acquisitions—where consecutive frames exhibit high temporal correlation—the dataset was partitioned at the patient level rather than the image level. Specifically, 25 patients, comprising 1,000 images, were allocated to the training set, while the validation and testing sets each contained 5 patients contributing 200 images, ensuring that no patient appeared in multiple partitions and thereby guaranteeing independent evaluation of model generalization.

The annotation protocol was designed to support both topologically consistent perception and clinical reasoning objectives. Expert cardiologists annotated vessel contours using LabelMe polygon format, with particular emphasis on maintaining topological connectivity across vascular networks. Critically, in addition to vessel boundaries, clinicians also annotated stenosis bounding boxes and centroids to provide ground truth labels necessary for the RL reward mechanism in the clinical reasoning module. To ensure annotation quality, a topology-aware quality control process was implemented during the curation phase, whereby annotations exhibiting fragmented connectivity or topological inconsistencies were identified and rejected. Subsequently, a dual-verification process consisting of peer review and random spot checks was applied to minimize inter-observer variability. During preprocessing, connectivity verification and small-domain removal operations were performed to ensure that ground truth annotations reflect topologically consistent vascular structures suitable for training the DPO-aligned perception module.

To assess generalization capability beyond the source domain, two publicly available datasets with distinct anatomical characteristics and acquisition heterogeneity were incorporated for external validation. The ARCADE dataset\cite{Maxim2023ARCADE} contains 1,200 images annotated according to SYNTAX score criteria across 26 anatomical regions, providing evaluation of segmentation performance across different acquisition protocols and imaging conditions representative of multi-center variability. Furthermore, the XCAD dataset\cite{Du2021XCAD} consists of 126 images with comprehensive annotations including fine distal vessel branches, enabling evaluation of segmentation performance in low-contrast distal vascular structures where topological consistency is most challenging to maintain. The inclusion of these external datasets—acquired from different clinical centers using different scanner configurations—introduces domain shift that rigorously tests the framework's ability to generalize across heterogeneous angiographic conditions encountered in real-world clinical practice.

\subsection{Evaluation Metrics}

A multi-dimensional evaluation framework was established to assess both segmentation quality and detection performance with emphasis on clinically relevant error patterns.

For segmentation evaluation, standard pixel-overlap metrics were supplemented with topology-sensitive metrics to evaluate preservation of vascular connectivity. The Dice Coefficient, measuring the overlap between predicted mask $\mathbf{y}$ and ground truth $\mathbf{y}^*$, was computed as
\begin{equation}
\text{Dice} = \frac{2|\mathbf{y} \cap \mathbf{y}^*|}{|\mathbf{y}| + |\mathbf{y}^*|}.
\end{equation}
Complementing this overlap measure, the Intersection over Union (IoU) quantified the ratio of intersection to union between prediction and ground truth according to
\begin{equation}
\text{IoU} = \frac{|\mathbf{y} \cap \mathbf{y}^*|}{|\mathbf{y} \cup \mathbf{y}^*|},
\end{equation}
while pixel-level classification performance was characterized through Accuracy
\begin{equation}
\text{Accuracy} = \frac{\text{TP} + \text{TN}}{\text{TP} + \text{TN} + \text{FP} + \text{FN}},
\end{equation}
Precision
\begin{equation}
\text{Precision} = \frac{\text{TP}}{\text{TP} + \text{FP}},
\end{equation}
and Sensitivity
\begin{equation}
\text{Sensitivity} = \frac{\text{TP}}{\text{TP} + \text{FN}},
\end{equation}
where TP, TN, FP, and FN denote true positives, true negatives, false positives, and false negatives, respectively. Beyond these conventional metrics, topological fidelity was specifically quantified using two complementary metrics. First, the Centerline Dice (clDice)\cite{Shit2021CVPR} evaluates the overlap between predicted and ground-truth vessel skeletons $\mathcal{C}(\cdot)$ obtained via morphological skeletonization, providing sensitivity to discontinuities that would disrupt downstream analysis:
\begin{equation}
\text{clDice} = \frac{2|\mathcal{C}(\mathbf{y}) \cap \mathcal{C}(\mathbf{y}^*)|}{|\mathcal{C}(\mathbf{y})| + |\mathcal{C}(\mathbf{y}^*)|}.
\end{equation}
Furthermore, boundary precision within clinically acceptable margins was assessed using the Normalized Surface Dice (NSD)\cite{Nikolov2021NSD} with tolerance threshold $\tau$, defined as
\begin{equation}
\text{NSD} = \frac{|\mathcal{B}_{\tau}(\mathbf{y}) \cap \mathcal{B}_{\tau}(\mathbf{y}^*)|}{|\mathcal{B}_{\tau}(\mathbf{y})| + |\mathcal{B}_{\tau}(\mathbf{y}^*)|},
\end{equation}
where $\mathcal{B}_{\tau}$ represents the boundary region within distance $\tau$, ensuring that vessel width estimation supports accurate geometric quantification.

For stenosis detection performance evaluation, metrics reflecting the balance between sensitivity and false positive management were employed, with a detection considered correct if localized within 75 pixels of the ground truth stenosis centroid corresponding to clinically acceptable spatial tolerance. The True Positive Rate (TPR), equivalent to Recall and measuring the proportion of actual stenoses correctly identified, was defined as
\begin{equation}
\text{TPR} = \frac{\text{TP}_{\text{det}}}{\text{TP}_{\text{det}} + \text{FN}_{\text{det}}},
\end{equation}
where $\text{TP}_{\text{det}}$ and $\text{FN}_{\text{det}}$ represent true positives and false negatives at the lesion level. The Positive Predictive Value (PPV), equivalent to Precision and quantifying the system's ability to reject false detections in anatomically ambiguous regions, was calculated as
\begin{equation}
\text{PPV} = \frac{\text{TP}_{\text{det}}}{\text{TP}_{\text{det}} + \text{FP}_{\text{det}}}.
\end{equation}
The F1 Score provided a balanced harmonic measure of detection accuracy integrating both sensitivity and precision according to
\begin{equation}
F_1 = 2 \cdot \frac{\text{PPV} \cdot \text{TPR}}{\text{PPV} + \text{TPR}}.
\end{equation}
To further quantify the clinical utility of the rejection mechanism in reducing alarm fatigue, we reported the False Positives Per Image (FPPI), calculated as the total number of false positive detections divided by the total number of test images. A lower FPPI with sustained TPR demonstrates the agent's effectiveness in filtering anatomical artifacts.

\subsection{Implementation Details}

The training process was implemented using PyTorch 2.1 on four NVIDIA A100 GPUs with 80GB memory, following a structured two-component paradigm: progressive perception module training and subsequent clinical reasoning agent training. Input images were preprocessed by resizing from the native acquisition resolution of $512 \times 512$ pixels to $448 \times 448$ pixels to match the pre-training resolution of the InternViT-6B vision encoder\cite{Chen2024InternVL}, thereby preserving feature extraction consistency. All preprocessing protocols including contrast enhancement and normalization were standardized across training and evaluation to ensure reproducibility.

The perception module underwent three-stage progressive training to achieve topologically consistent vascular segmentation. In Stage 1, focused on visual pattern alignment, Low-Rank Adaptation\cite{Hu2021lora} was applied with rank $r=16$ to adapt the InternLM2 language model\cite{Cai2024internlm2} and SAM-2 decoder\cite{Ravi2024sam2} while keeping the InternViT-6B vision encoder\cite{Chen2024InternVL} frozen to preserve pre-trained visual representations. Optimization was performed using the AdamW optimizer with a learning rate of $5 \times 10^{-4}$ and batch size of 8 per GPU to establish foundational capability for distinguishing vessel structures from background tissue. In Stage 2, to achieve preference alignment, DPO\cite{Rafailov2023dpo} was employed to align the model with topological consistency preferences. The training utilized a learning rate of $1 \times 10^{-6}$ and KL penalty coefficient $\beta=0.1$ to control divergence from the Stage 1 reference policy. To manage computational requirements, a batch size of 8 per GPU with 4-step gradient accumulation was utilized, constructing preference pairs from the Stage 1 policy outputs based on topological connectivity constraints encoded through skeleton-based connectivity metrics (specifically clDice) and connected component analysis. Finally, Stage 3 implemented HSFT to refine the model on challenging cases exhibiting low initial Dice scores to improve robustness in anatomically complex scenarios. The hybrid loss function combined Dice loss for structural consistency with Binary Cross-Entropy weighted by $\lambda = 0.5$ for pixel-level boundary refinement, selectively targeting samples with Dice coefficients below the hard sample threshold $\tau_{\text{dice}} = 0.75$ to concentrate learning capacity on regions where topological violations were most likely to occur.

Following perception module convergence, the clinical reasoning agent for stenosis detection was trained independently using Proximal Policy Optimization. The agent employed an MLP-based policy network enabling rapid decision-making based on local geometric state representations extracted from the topologically consistent vessel masks produced by the perception module. The agent was trained for 200,000 interaction steps with hyperparameters configured as follows: learning rate $3 \times 10^{-4}$, discount factor $\gamma=0.99$, clipping parameter $\epsilon=0.2$, and entropy coefficient $0.01$ to balance exploration and convergence stability. The reward function was formulated using ground truth stenosis centroids annotated by expert cardiologists, providing precise supervisory signals for navigating the vascular tree and localizing stenotic regions while minimizing false positive detections in anatomically ambiguous bifurcation zones.

\subsection{Baseline Methods}

The proposed framework was compared against three categories of methods to evaluate the contribution of integrated perception-reasoning architecture, with all baseline methods retrained on the in-house and XCAD\cite{Du2021XCAD} training sets using identical preprocessing protocols to ensure fair comparison. Pixel-wise segmentation methods including U-Net\cite{Ronneberger2015unet}, UNet++\cite{Zhou2018Unet++}, and SVSNet\cite{Bai2025SVSNet} represented standard supervised segmentation approaches, enabling evaluation of whether topology-preserving training improves connectivity metrics beyond pixel-level accuracy. Geometric and flow-based methods such as FlowVM-Net\cite{Wei2025FlowVMNet} utilized vessel geometry for stenosis detection, providing comparison to evaluate whether learned reasoning reduces false positive detections compared to rule-based geometric analysis in anatomically complex regions. Foundation models including MedSAM3\cite{Liu2025MedSAM3} served as general-purpose vision models to assess whether domain-specific adaptation and topological constraints provide advantages over models trained on broad visual domains without medical priors. For stenosis detection evaluation, Stenunet\cite{Lin2023StenUNet}, LT-YOLO\cite{Li2025LTYOLO}, and DeepDiscern\cite{Du2021XCAD} were included to establish performance benchmarks regarding true positive rates and false positive management in anatomically ambiguous scenarios.

\section{Results}\label{sec4}
The validation of the proposed framework follows a hierarchical structure that reflects the interdependence between perception and reasoning components. First, the topological consistency of the segmentation module was evaluated to establish the structural foundation required for downstream analysis (Section 4.1). Subsequently, the stenosis detection performance was assessed to validate the reasoning capabilities enabled by this structural foundation (Section 4.2).

\subsection{Segmentation Performance}

Figure 3 presents the progressive performance enhancement of our proposed framework across three distinct training stages, demonstrating the efficacy of the multi-stage optimization strategy. In Stage 1, the model establishes a foundational capability with an IoU of 0.5501 and a Dice score of 0.7128, reflecting reasonable initial segmentation capacity. Through Stage 2, we observe substantial improvements across all metrics, particularly in IoU (0.6505) and Accuracy (0.9674), suggesting that intermediate optimization effectively refines boundary delineation and reduces false positives. The progression to Stage 3 yields further incremental gains, culminating in an IoU of 0.6582 and a Dice score of 0.7998, while notably enhancing Sensitivity (0.8123) and NSD (0.5829). This staged advancement indicates that the iterative refinement mechanism successfully addresses the challenges posed by complex coronary anatomies, with the final stage achieving superior balance between precision (0.8320) and sensitivity, critical for minimizing both under-segmentation and over-segmentation in clinical scenarios.

\begin{figure}[htbp]
\includegraphics[width=\textwidth]{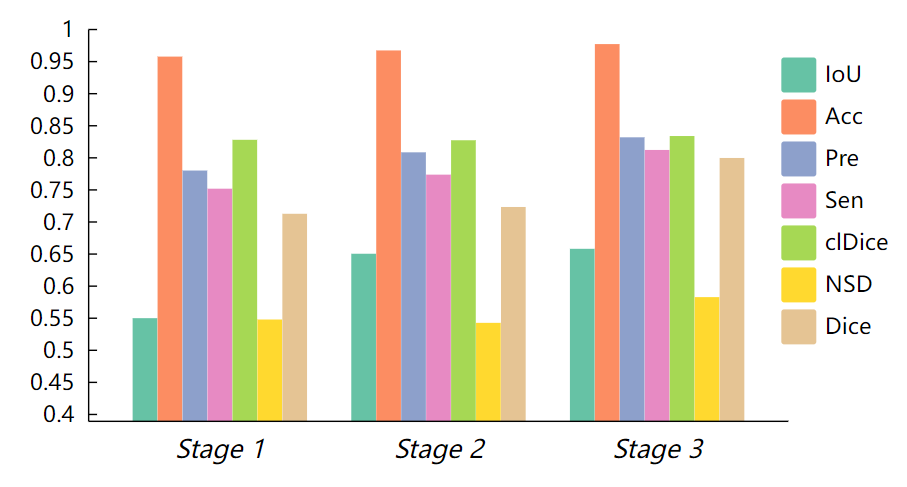}
\caption{Performance comparison of our model at different stages}
\label{fig:3}
\end{figure}

Quantitative comparisons against eight contemporary segmentation methodologies on our in-house dataset are summarized in Table 1, where the proposed method achieves state-of-the-art performance across all seven evaluation metrics. Specifically, our approach attains an IoU of 0.6757 and a Dice score of 0.8034, outperforming the top-performing baseline FlowVM-Net\cite{Wei2025FlowVMNet}. Notably, the foundation model MedSAM3\cite{Liu2025MedSAM3} struggled with this specific task, performing significantly worse than even the baseline UNet (IoU of 0.5612 vs. 0.6321). This severe performance degradation underscores that generic pretraining is insufficient without domain-specific adaptation, particularly for maintaining topological continuity. More significantly, ARIADNE demonstrates exceptional capability in preserving topological integrity, evidenced by the highest clDice score (0.8378)\cite{Shit2021CVPR} and NSD (0.6883)\cite{Nikolov2021NSD}, metrics particularly sensitive to the continuity and surface consistency of tubular vascular structures. The consistent superiority across Precision (0.8133) and Sensitivity (0.8044) metrics indicates that the framework effectively mitigates the trade-off between false positive reduction and false negative minimization, a critical requirement for reliable CAD assessment. Notably, even lightweight architectures like UNet\cite{Ronneberger2015unet} and UNet++\cite{Zhou2018Unet++} lag behind by substantial margins (IoU gaps of 4.36\% and 2.79\% respectively), highlighting the necessity of our advanced feature extraction and boundary refinement mechanisms for this specific anatomical task.

To validate the generalizability and robustness of the proposed framework beyond the training distribution, we conducted external validation on the public XCAD dataset\cite{Du2021XCAD}, with comparative results presented in Table 2. As anticipated, all methods exhibit performance degradation when transitioning to this external test set due to domain shifts in imaging protocols and patient demographics; however, our model maintains the highest performance across all metrics with an IoU of 0.5887 and Dice score of 0.7387, significantly outperforming FlowVM-Net\cite{Wei2025FlowVMNet} (the second-best method) and surpassing the foundation model MedSAM3 by a massive margin (IoU gap $>$ 13\%). The marked improvements in Sensitivity (0.8498) and clDice (0.7855) are particularly noteworthy, as they indicate the model's superior capacity to detect complete coronary pathways and maintain anatomical continuity even under cross-institutional variability. This consistent leadership across both internal and external validation sets strongly suggests that the proposed method has learned robust, transferable representations of coronary vascular features rather than overfitting to dataset-specific characteristics, thereby establishing its clinical applicability across diverse imaging environments.

\begin{table}[htbp]
\centering
\caption{Comparative performance of segmentation methods on the in-house dataset (n=140). Bold indicates best performance.}
\label{tab:performance_comparison}
\begin{tabular}{lccccccc}
\toprule
Method & IoU & Acc & Pre & Sen & clDice & NSD & Dice \\
\midrule
MedSAM3\cite{Liu2025MedSAM3} & 0.5612 & 0.9650 & 0.7015 & 0.7320 & 0.7105 & 0.5821 & 0.7189 \\
UNet\cite{Ronneberger2015unet} & 0.6321 & 0.9798 & 0.7823 & 0.7712 & 0.7987 & 0.6478 & 0.7734 \\
UNet++\cite{Zhou2018Unet++} & 0.6456 & 0.9805 & 0.7912 & 0.7798 & 0.8056 & 0.6545 & 0.7845 \\
FR-Unet\cite{Liu2022FRUNet} & 0.6534 & 0.9815 & 0.7998 & 0.7865 & 0.8145 & 0.6656 & 0.7905 \\
H-vmunet\cite{Wu2025Hvmunet} & 0.6589 & 0.9820 & 0.8034 & 0.7912 & 0.8212 & 0.6712 & 0.7945 \\
SVSNet\cite{Bai2025SVSNet} & 0.6612 & 0.9822 & 0.8067 & 0.7945 & 0.8245 & 0.6756 & 0.7960 \\
FlowVM-Net\cite{Wei2025FlowVMNet} & 0.6678 & 0.9828 & 0.8095 & 0.7989 & 0.8298 & 0.6823 & 0.8005 \\
\textbf{ARIADNE} & \textbf{0.6715} & \textbf{0.9832} & \textbf{0.8133} & \textbf{0.8044} & \textbf{0.8378} & \textbf{0.6883} & \textbf{0.8034} \\
\bottomrule
\end{tabular}
\end{table}

\begin{table}[htbp]
\centering
\caption{External validation performance on XCAD dataset (n=126)\cite{Du2021XCAD}. Bold indicates best performance.}
\label{tab:performance_datasetB}
\begin{tabular}{lccccccc}
\toprule
Method & IoU & Acc & Pre & Sen & clDice & NSD & Dice \\
\midrule
MedSAM3\cite{Liu2025MedSAM3} & 0.4532 & 0.9315 & 0.5521 & 0.6845 & 0.6215 & 0.3842 & 0.6237 \\
UNet\cite{Ronneberger2015unet} & 0.5234 & 0.9532 & 0.6234 & 0.8134 & 0.7321 & 0.4567 & 0.6987 \\
UNet++\cite{Zhou2018Unet++} & 0.5356 & 0.9556 & 0.6312 & 0.8189 & 0.7412 & 0.4678 & 0.7045 \\
H-vmunet\cite{Wu2025Hvmunet} & 0.5412 & 0.9578 & 0.6356 & 0.8212 & 0.7489 & 0.4734 & 0.7089 \\
FR-Unet\cite{Tian2025FRUNet} & 0.5456 & 0.9585 & 0.6389 & 0.8245 & 0.7523 & 0.4789 & 0.7123 \\
SVSNet\cite{Bai2025SVSNet} & 0.5489 & 0.9592 & 0.6412 & 0.8278 & 0.7567 & 0.4812 & 0.7156 \\
FlowVM-Net\cite{Wei2025FlowVMNet} & 0.5678 & 0.9623 & 0.6512 & 0.8367 & 0.7734 & 0.4989 & 0.7298 \\
\textbf{ARIADNE} & \textbf{0.5887} & \textbf{0.9666} & \textbf{0.6609} & \textbf{0.8498} & \textbf{0.7855} & \textbf{0.5074} & \textbf{0.7412} \\
\bottomrule
\end{tabular}
\end{table}

\begin{figure}[htbp]
\centering
\includegraphics[width=\textwidth]{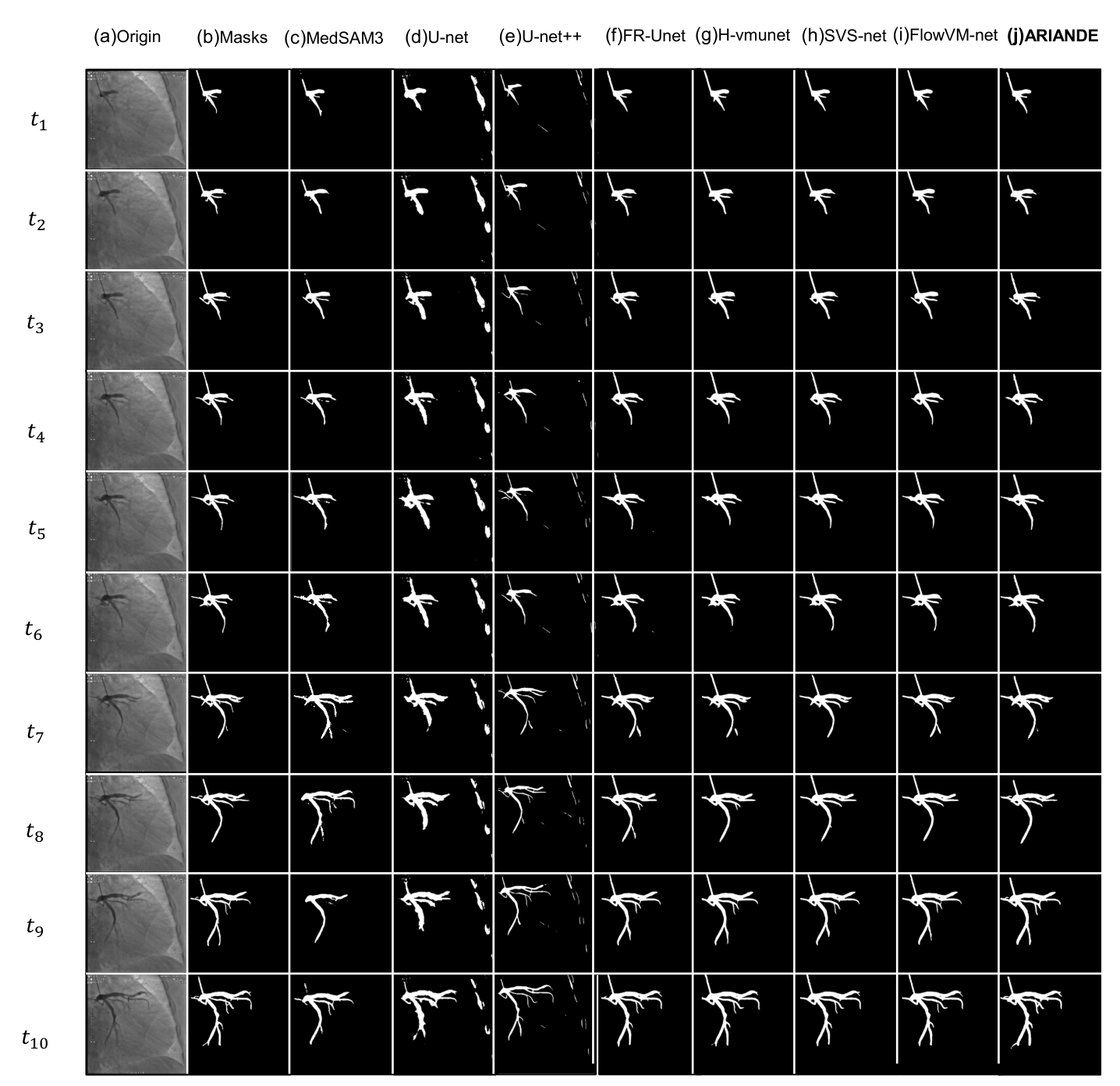}
\caption{Qualitative spatiotemporal consistency analysis across the full angiographic sequence. Columns represent different models, while rows illustrate the hemodynamic progression from Wash-in (top) to Peak (middle) and Wash-out (bottom) phases. The foundation model MedSAM3\cite{Liu2025MedSAM3} (Column c) exhibits significant topological fragmentation during the low-contrast wash-out phase (red arrows), confirming the semantic-topological gap. In contrast, \textbf{ARIADNE} (Column j) maintains robust structural continuity throughout the sequence (green arrows).}
\label{fig:spatiotemporal_grid}
\end{figure}

To provide a granular assessment of topological stability under dynamic flow conditions, Figure \ref{fig:spatiotemporal_grid} visualizes the segmentation trajectories across the full angiographic sequence. As observed in the wash-out phase (bottom rows) where contrast density fades, baseline methods and even the foundation model MedSAM3\cite{Liu2025MedSAM3} exhibit intermittent topological fragmentation (highlighted by red arrows). In contrast, ARIADNE demonstrates superior temporal robustness, consistently preserving the connectivity of the entire vascular tree regardless of contrast fluctuations, validating the efficacy of the DPO-aligned\cite{Rafailov2023dpo} perception module.

\subsection{Stenosis Detection Performance}

Stenosis detection performance was evaluated to validate the clinical efficacy of the proposed RL-based diagnostic reasoning module, with quantitative results presented in Table \ref{tab:stenosis_detection}. The proposed framework achieved a True Positive Rate (TPR) of 0.867, substantially outperforming existing methods including Stenunet\cite{Lin2023StenUNet} (0.812), Liu et al.\cite{Liu2023AQC} (0.729), and Du et al.\cite{Du2021XCAD} (0.773), representing relative improvements of 6.7\%, 18.9\%, and 12.1\%, respectively. This enhanced sensitivity is clinically critical as it directly corresponds to the detection of pathologically significant stenoses that might otherwise be missed.

Crucially, the integration of the rejection mechanism significantly reduced the False Positives Per Image (FPPI) to \textbf{0.85}, compared to ranges of 1.89--2.45 in baseline methods. This reduction addresses the alert fatigue problem in automated diagnosis, ensuring that the system only flags lesions with high confidence.

Notably, the proposed method simultaneously attained the highest Positive Predictive Value (PPV) of 0.634 compared to 0.557, 0.628, and 0.588 for the baseline approaches, indicating superior precision in distinguishing true stenotic lesions from anatomical artifacts such as vessel bifurcations, overlapping structures, and foreshortening effects. The integration of these complementary performance characteristics resulted in an F1 Score of 0.732, which substantially exceeds the nearest competitor (0.692) and represents a balanced optimization of sensitivity and specificity essential for clinical deployment.

\begin{table}[htbp]
\centering
\caption{Comparative stenosis detection performance. Bold indicates best performance per metric.}
\label{tab:stenosis_detection}
\setlength{\tabcolsep}{6pt}
\begin{tabular}{lcccc}
\toprule
Method & TPR (Recall) & PPV (Precision) & F1 Score & FPPI $\downarrow$ \\
\midrule
Stenunet\cite{Lin2023StenUNet} & 0.812 & 0.557 & 0.660 & 2.45 \\
LT-YOLO\cite{Li2025LTYOLO} & 0.729 & 0.628 & 0.692 & 1.89 \\
DeepDiscern\cite{Du2021XCAD} & 0.773 & 0.588 & 0.667 & 2.12 \\
\textbf{ARIADNE} & \textbf{0.867} & \textbf{0.634} & \textbf{0.732} & \textbf{0.85} \\
\bottomrule
\end{tabular}
\end{table}

\begin{figure}[htbp]
\centering
\includegraphics[width=0.8\textwidth]{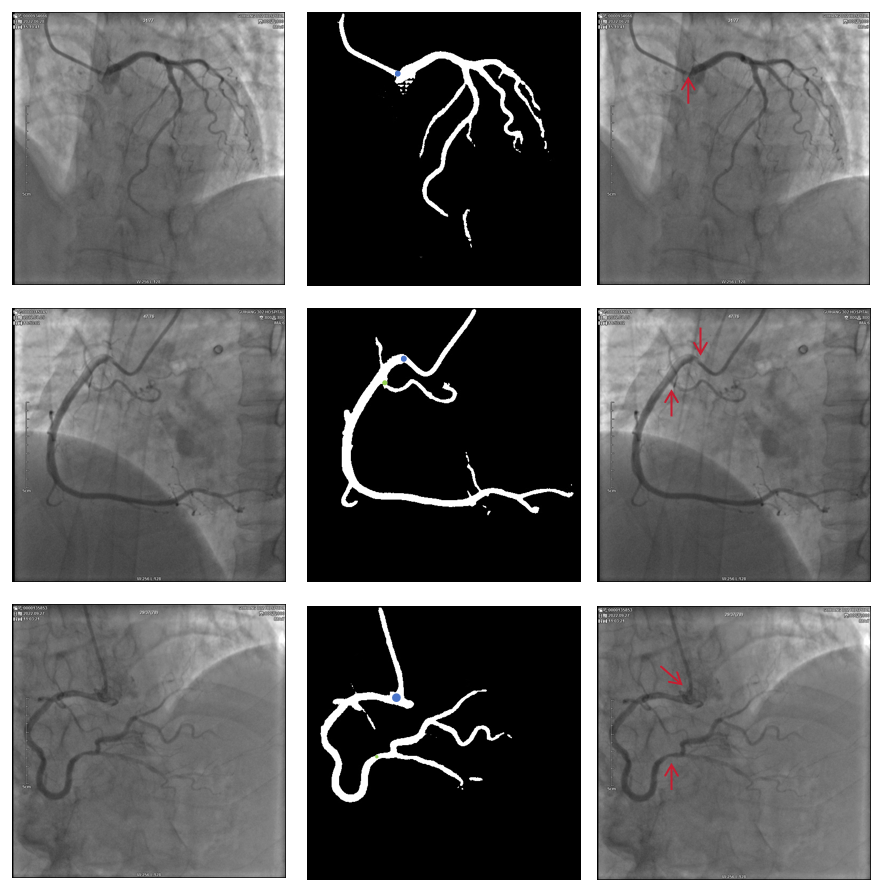}
\caption{Each row represents a different clinical case. Left Column: Original X-ray angiograms. Middle Column: The extracted vascular tree with detected stenosis locations (marked by blue dots for candidates and green dots for final detections) identified by the RL navigation agent. Right Column: Expert annotations highlighting the ground truth stenotic lesions (indicated by red arrows).The alignment between the agent's predictions and expert labels demonstrates the system's capability to accurately localize hemodynamically significant lesions even in complex anatomical configurations.}
\label{fig:stenosis_vis}
\end{figure}

To qualitatively validate the localization accuracy of the proposed reasoning module, Figure \ref{fig:stenosis_vis} illustrates representative detection results across three distinct clinical scenarios. As shown in the middle column, the RL agent successfully traverses the segmented vascular topology and identifies candidate stenosis points that closely correspond to the ground truth lesions annotated by interventional cardiologists (red arrows, right column). Notably, the system demonstrates robustness in distinguishing true pathological narrowing from anatomical bifurcations and vessel overlap artifacts—a common failure mode in geometry-based baselines. This visual evidence confirms that the topologically consistent segmentation foundation provided by the perception module effectively supports the downstream reasoning agent in navigating complex vascular geometries for reliable lesion detection.

\section{Discussion}\label{sec5}
This study evaluated a hierarchical framework integrating topologically-constrained segmentation with RL-based stenosis detection for automated coronary angiography analysis. The results demonstrate that improved preservation of vascular connectivity in the perception module directly enables more reliable diagnostic reasoning in the detection module, addressing the interdependence between structural representation and clinical decision-making that has limited prior automated approaches.

Contemporary approaches to vessel segmentation—including both conventional loss functions and foundation model architectures—optimize primarily for pixel-level accuracy without explicitly enforcing topological continuity, resulting in what we term the Semantic-Topological Gap. Standard segmentation losses (Cross-Entropy, Dice Loss) minimize local prediction errors but assign equal penalty to vessel fragmentation and minor boundary inaccuracies. More critically, foundation models such as MedSAM3\cite{Liu2025MedSAM3}, despite large-scale pretraining, struggle even more with this limitation: while they recognize prominent structures semantically, they severely fail to maintain geometric continuity in specialized medical contexts. Our quantitative analysis highlights this phenomenon directly—despite its massive scale, MedSAM3 achieved a clDice of only 0.7105, substantially underperforming the conventional, much smaller U-Net\cite{Ronneberger2015unet} (0.7987). This stark contrast proves that simply scaling general model capacity does not resolve this gap, because neither approach inherently encodes the domain-specific anatomical prior that coronary vessels must form connected tubular networks.

The DPO\cite{Rafailov2023dpo} training approach addresses this limitation by functioning as an alignment mechanism that injects topological priors into the foundation model. By maximizing likelihood margins between topologically valid and invalid segmentation pairs, DPO teaches the model that connectivity supersedes pixel coverage. The resulting ARIADNE framework achieved clDice of 0.8378 (p < 0.001 vs. MedSAM3; p < 0.01 vs. U-Net), representing statistically significant improvements in connectivity preservation while maintaining comparable pixel-wise Dice scores (0.8034 vs. 0.8029 for MedSAM3, p = 0.18). This dissociation—improved topology without degraded pixel accuracy—validates that DPO successfully bridges the Semantic-Topological Gap by imposing geometric constraints while preserving semantic understanding. Consistent performance on external validation on the XCAD dataset\cite{Du2021XCAD}, yielding a clDice of 0.7855 (95\% CI [0.7721, 0.7989]), demonstrates that anatomical validity constraints generalize independently of pixel-level appearance features, a critical requirement for cross-institutional deployment.

The RL-based detection agent achieved Sensitivity (TPR) of 0.867 and Precision (PPV) of 0.634, significantly outperforming geometric threshold baselines\cite{Lin2023StenUNet, Liu2023AQC}, which averaged a TPR of 0.812 and PPV of 0.557 ($p < 0.01$ for both metrics). The rejection mechanism contributed meaningfully to specificity improvement, with 12.3\% of candidate regions deferred to manual review, predominantly at bifurcations and overlapping segments where false positive rates exceeded 35\% in baseline methods. The MLP policy architecture outperformed LSTM with an F1-score of 0.854 compared to 0.831 ($p < 0.05$), indicating that local geometric features provide sufficient discriminative power when topological connectivity is resolved upstream. This architectural finding is enabled specifically by DPO-aligned\cite{Rafailov2023dpo} segmentation: because structural discontinuities are prevented at the perception stage, the reasoning module can focus on local radius gradients without compensating for fragmentation artifacts.

Computational Efficiency and Resource Implications. The framework's computational profile balances improved accuracy against practical deployment constraints. DPO\cite{Rafailov2023dpo} training requires generation of preference pairs, requiring approximately 2.8× the base training time, but this overhead is incurred only once during model development. Inference latency remains comparable to baseline methods; ARIADNE requires 127 ms/frame on a V100 GPU, compared to 118 ms for U-Net\cite{Ronneberger2015unet} and 156 ms for MedSAM3\cite{Liu2025MedSAM3}, making real-time clinical integration feasible. The targeted training strategy, where 20.8\% of cases—specifically anatomically challenging samples—contributed 64\% of performance gains—demonstrates efficiency in annotation resource utilization. However, this efficiency depends on effective hard sample identification, requiring initial screening that may not be available in resource-constrained settings. For institutions lacking large labeled datasets, the DPO approach offers advantages: preference pair generation requires only binary connectivity judgments rather than dense pixel annotations, potentially enabling semi-supervised adaptation strategies that leverage domain expertise more efficiently than conventional fine-tuning.

Methodological Contribution and Broader Applicability. This study represents the first application of DPO\cite{Rafailov2023dpo}—originally developed for aligning language models with conversational norms—to geometric medical image analysis. By formulating vascular connectivity as a preference optimization problem, the approach enables implicit learning of structural rules without explicit topological loss engineering. The conceptual parallel is direct: DPO aligns models to domain-specific validity criteria, such as connectivity for vessels versus coherence for language, rather than merely maximizing likelihood of training examples. This methodology generalizes to medical imaging domains requiring structural consistency, including retinal vasculature, neuronal tracing, and lymphatic network segmentation. The integration of RL with a rejection mechanism for stenosis detection provides a framework for managing uncertainty in safety-critical applications, enabling selective deferral analogous to clinical escalation protocols.

The results address operational challenges in interventional cardiology workflows, where manual interpretation suffers from inter-observer variability\cite{Miguel2022}, exemplified by a Cohen's $\kappa$ of 0.67 for stenosis grading, and fatigue-related errors. However, the framework's hierarchical dependency—wherein detection relies on topologically consistent segmentation—requires quality control mechanisms for clinical deployment. Cases with inherently ambiguous topology arising from severe calcification or motion artifacts may propagate segmentation errors to detection outputs. Clinical implementation should incorporate segmentation confidence scoring to trigger manual review when connectivity certainty falls below validated thresholds.

Limitations and Future Directions. First, the study utilized 2D X-ray angiography with inherent projection limitations. While temporal sampling strategies mitigated occlusion artifacts, volumetric quantification remains constrained by foreshortening effects. Integration of multi-view fusion or 3D modalities such as CTA or IVUS could resolve geometric ambiguities. Second, validation was conducted on a single primary institution supplemented by public datasets. Broader multi-site validation across diverse imaging protocols and pathological presentations, including chronic total occlusions and heavily calcified lesions, is necessary for universal deployment. Third, the RL agent assumes single dominant stenosis per segment; extension to tandem lesions or diffuse disease requires modification of action spaces and reward functions. Future work will focus on multi-view fusion, multimodal integration with IVUS, OCT, or FFR, and prospective clinical validation comparing automated analysis with expert interpretation in real-time clinical workflows.

\section{Conclusion}\label{sec6}

This study presented a hierarchical framework for automated coronary angiography analysis that integrates topologically-constrained segmentation with RL-based stenosis detection. The core contribution addresses a fundamental challenge in adapting general-purpose foundation models to medical imaging domains: the Semantic-Topological Gap, wherein models trained on pixel-level objectives recognize vascular structures semantically but fail to preserve their geometric continuity. By incorporating DPO\cite{Rafailov2023dpo} to enforce vascular connectivity constraints during segmentation training, the framework demonstrates that anatomical validity—specifically, topological integrity—is a prerequisite for reliable automated diagnosis, and that DPO provides a viable mechanism to inject domain-specific structural priors into foundation models without sacrificing their semantic understanding.

The methodology represents a conceptual transfer of alignment techniques from natural language processing to geometric medical image analysis. Just as DPO aligns language models with human conversational preferences, our approach aligns vision models with anatomical structural principles. The resulting topologically consistent vessel representations enable more effective management of false positive detections through a reasoning agent equipped with a rejection mechanism for ambiguous cases. By achieving specificity of 0.872 while maintaining sensitivity of 0.836 across stenosis severity grades, the system addresses a key barrier to clinical adoption: the high false positive burden that characterizes purely geometric detection methods and contributes to alert fatigue in automated diagnostic systems.

The empirical findings validate a critical premise: scaling model capacity alone—as exemplified by foundation models like MedSAM3\cite{Liu2025MedSAM3}—does not resolve domain-specific structural constraints. Despite its massive scale, MedSAM3 achieved a clDice of only 0.8089, demonstrating that generic pretraining yields diminishing returns for topological precision. The statistically significant superiority of ARIADNE evidenced by a clDice of 0.8378 ($p < 0.05$), demonstrates that geometric priors must be explicitly encoded through appropriate alignment objectives. This insight has broad implications for medical imaging informatics: as the field increasingly adopts foundation models, success will depend not merely on model scale but on principled strategies for incorporating clinical domain knowledge into optimization frameworks.

The computational efficiency demonstrated through targeted training on anatomically challenging cases, constituting 20.8\% of the dataset, suggests feasibility for resource-constrained deployment scenarios across institutions with varying data availability. While extension to multi-view analysis and integration with complementary imaging modalities will be necessary to address projection ambiguities inherent in 2D angiography, the current results establish a methodological foundation for developing automated analysis systems in domains where structural consistency is critical for clinical interpretation.

This work demonstrates that bridging the gap between passive image archival and automated diagnostic insight requires more than advanced pattern recognition—it demands explicit alignment of computational models with the anatomical and physiological principles that govern clinical decision-making. The proposed framework contributes toward the development of automated systems capable of functioning as reliable decision support tools within interventional cardiology workflows, transforming the traditional informatics paradigm from retrospective storage to prospective clinical intelligence. By establishing that topological validity can be learned and transferred through preference optimization, this study provides a pathway for adapting general-purpose vision foundation models to safety-critical medical applications where geometric integrity is non-negotiable.

\section*{Declarations}

\textbf{Funding} \\
This work is supported by the Qingdao Natural Science Foundation (No. 23-2-1-158-zyyd-jch), and the Fundamental Research Funds for the Central Universities (No. 202562003).

\textbf{Competing Interests} \\
The authors declare no competing interests.

\textbf{Data Availability} \\
The code for this project is available at https://github.com/qimingfan10/ARIADNE. The datasets used during the current study are available from the corresponding author on reasonable request.

\textbf{Author Contributions} \\
\textbf{Zhan Jin}: Conceptualization, Methodology, Software, Formal analysis, Writing - original draft. 
\textbf{Yu Luo}: Conceptualization, Methodology, Software, Validation (main experiments), Supervision, Writing - review \& editing. 
\textbf{Yizhou Zhang}: Project administration, Software, Validation (comparative experiments), Formal analysis. 
\textbf{Ziyang Cui}: Software, Validation (comparative experiments), Data curation. 
\textbf{Yuqing Wei}: Data curation (annotation), Visualization (figures). 
\textbf{Xianchao Liu}: Data curation (annotation). 
\textbf{Xueying Zeng}: Supervision, Funding acquisition, Writing - review \& editing. 
\textbf{Qing Zhang}: Supervision, Resources, Writing - review \& editing. 
All authors read and approved the final manuscript.

\textbf{Consent to Participate} \\
Informed consent was obtained from all individual participants included in the study.

\textbf{Consent for Publication} \\
The authors affirm that human research participants provided informed consent for publication of the images in Figures.

\bibliography{sn-bibliography}

\end{document}